\documentclass[runningheads]{llncs}
\usepackage[T1]{fontenc}

\usepackage{graphicx}
\usepackage{amsmath}
\usepackage{xcolor}
\usepackage{multirow}

\usepackage[hidelinks]{hyperref}
\usepackage{orcidlink}

\usepackage{microtype}

\begin{document}

\title{Towards Real-Time PixOOD: Efficient Anomaly Segmentation for Autonomous Vehicles}

\titlerunning{Towards Real-Time PixOOD}

\author{Luca de Martino\,\orcidlink{0009-0008-3487-6090} \and
Federico Aromolo\,\orcidlink{0009-0007-3537-8782} \and
Federico Nesti\,\orcidlink{0000-0003-4338-9573} \and
Giorgio Buttazzo\,\orcidlink{0000-0003-4959-4017}}

\authorrunning{L. de Martino et al.}

\institute{Department of Excellence in Robotics \& AI, Scuola Superiore Sant’Anna, Pisa, Italy}

\maketitle

\begin{abstract}

Real-time anomaly segmentation is essential for the safety of autonomous systems. Although recent approaches offer high accuracy, their computational cost limits their deployment on embedded hardware. This work presents an efficient and accelerated pipeline designed for both embedded and desktop platforms, targeting the autonomous driving and railway domains.
The proposed approach reformulates the Neyman--Pearson scoring stage of PixOOD, a state-of-the-art out-of-distribution detection method, and deploys the full pipeline through hardware-optimized TensorRT compilation, reaching up to 182\,FPS on a desktop NVIDIA RTX~4060 GPU and 75\,FPS on the NVIDIA Jetson AGX Orin embedded platform, respectively $20\times$ and $18\times$ faster than the original baseline. The achieved results demonstrate that advanced anomaly segmentation can be efficiently deployed for onboard processing in autonomous driving and railway applications.

\end{abstract}

\section{Introduction} \label{sec:intro}

In autonomous driving and railway applications, the ability to identify previously unseen objects at the pixel level, such as obstacles on the tracks or unexpected debris on the roadway, is essential for safe operation~\cite{LaF,nesti2026osdar}. Anomaly segmentation addresses this challenge by producing a dense per-pixel score map that flags pixels deviating from the learned distribution, thus enabling downstream safety modules to trigger appropriate responses. Performing such segmentation in real time is therefore crucial, as autonomous driving systems must respond faster than a human driver~\cite{lin2018architectural}; as a result, this work targets 25\,FPS.

The research community has primarily focused on providing anomaly segmentation methods based on different sensor inputs~\cite{shoeb2025ood,bogdoll2022anomaly} and on establishing benchmarks that measure anomaly segmentation performance metrics, such as Average Precision (AP) and false positive rate at 95\,\% recall (FPR95). However, the characterization of real-time deployment on embedded hardware remains largely unexplored.
The railway domain has received even less attention in this regard and, to the best of our knowledge, OSDaR-AR~\cite{nesti2026osdar} represents the only publicly available benchmark for autonomous railway operations that provides pixel-level anomaly annotations, along with LiDAR point clouds calibrated and synchronized with RGB frames.

Among the families of anomaly segmentation methods, prototype-based approaches model the in-distribution manifold and flag statistical deviations as anomalous, without requiring explicit outlier exposure during training~\cite{shoeb2025ood}. A representative method of this family is PixOOD~\cite{vojir2024pixood}, ranked first, at the time of writing, in the LostAndFound (LaF)~\cite{LaF} benchmark for autonomous driving~\cite{shoeb2025ood}.
However, the original PixOOD implementation relies on DINOv2~\cite{oquab2023dinov2}, a Vision Transformer (ViT) backbone, in its largest version, ViT-L. Since DINOv2 is available in several model sizes, this work adopts the smaller ViT-S version, which is better suited to embedded systems, and applies the same choice to the more recent DINOv3~\cite{simeoni2025dinov3}. Even with this lighter backbone and a reduced input resolution of 640\,px, the embedded target reaches only 4.26\,FPS, far below the 25\,FPS requirement, because the CPU scoring stage dominates the per-frame latency. Specifically, the Neyman--Pearson scoring stage runs entirely on the CPU in the original implementation, preventing real-time operation on the embedded target.
These findings highlight two compounding limitations: (i) state-of-the-art anomaly segmentation methods do not leverage the hardware acceleration technologies necessary to meet real-time requirements, and (ii) standard benchmarks only evaluate detection accuracy without reporting deployment-relevant latency or energy consumption.

\noindent\textbf{Paper contributions.} To overcome these limitations, this paper provides three main contributions targeting the efficient deployment of PixOOD:
\begin{itemize}
    \item A GPU reimplementation of the Neyman--Pearson scoring stage that evaluates the per-class density in log-space and replaces the CPU cumulative-distribution lookup with a GPU-native operator, eliminating the per-frame CPU round-trip; reducing the per-frame latency at 640\,px by $3.78\times$ on the NVIDIA Jetson AGX Orin embedded GPU platform and by $4.66\times$ on the NVIDIA RTX~4060 desktop GPU, with less than 0.15\,\% AP deviation from the original implementation.
    \item A hardware-aware deployment pipeline using TensorRT (a high-performance inference framework by NVIDIA) compilation via an ONNX-to-TensorRT workflow (ONNX is an interoperable format for neural network models) on both the Jetson AGX Orin and the desktop NVIDIA RTX~4060 GPU, with FP16 and FP32 precision selected per model architecture, reaching up to 182\,FPS on the desktop GPU and 75\,FPS on the embedded platform, up to $20\times$ over the original baseline.
    \item A systematic cross-domain characterization of detection accuracy, latency, and energy on an embedded and a desktop platform, across the autonomous driving and railway domains.

\end{itemize}

\noindent\textbf{Paper organization.} The remainder of this paper is organized as follows: Section~\ref{sec:related} reviews the relevant related work; Section~\ref{sec:background} details the background and system model; Section~\ref{sec:method} describes the proposed optimizations; Section~\ref{sec:experimental} reports the experimental evaluation; and Section~\ref{sec:conclusions} presents the conclusions and future directions.

\section{Related work} \label{sec:related}

Building on the categorization of Shoeb et al.~\cite{shoeb2025ood}, this work groups anomaly segmentation approaches into the following five families.

\noindent\textbf{Uncertainty-based methods.} These approaches derive a scalar anomaly score at test time from the output of a semantic segmentation network, without modifying its architecture. Representative scoring functions include the standardized max logit~\cite{jung2021standardizedmaxlogitssimple} and the energy score~\cite{liu2020energy}. To improve discriminability, several state-of-the-art methods in this family additionally fine-tune the network with auxiliary outlier exposure, typically the COCO dataset~\cite{lin2014microsoft,shoeb2025ood}.

\noindent\textbf{Mask-based methods.} These methods rely on a mask-transformer decoder that produces a set of mask queries, each associated with a binary mask and a class distribution. A pixel is flagged as anomalous when no query of a known class claims it~\cite{nayal2023rbasegmentingunknownregions}. The computational cost of the query-based decoder limits the deployment of these methods on real-time embedded systems~\cite{shoeb2025ood}.

\noindent\textbf{Reconstructive methods.} These methods learn to reconstruct the input from a closed-set prediction and compute the anomaly score from the discrepancy between the original input and its reconstruction~\cite{lis2019detectingunexpectedimageresynthesis}. They do not require outlier exposure but are prone to false positives in complex or repetitive textures~\cite{shoeb2025ood}.

\noindent\textbf{Foundation model approaches.} These methods adapt large-scale vision foundation models to segment out-of-distribution objects~\cite{zhao2024segmentoutofdistributionobject}. However, the high computational cost of running such large models on embedded hardware prevents their real-time deployment~\cite{shoeb2025ood}.

\noindent\textbf{Prototype and density-based methods.} These methods parameterize in-distribution class manifolds and detect anomalies as statistical deviations from these models~\cite{vojir2024pixood}. Since these approaches rely exclusively on a model of in-distribution normality and require no anomalous data, they can be applied to new domains by retraining on in-distribution data alone.

\medskip
\noindent Among the methods described above, PixOOD~\cite{vojir2024pixood}, a prototype-based method, achieves state-of-the-art accuracy on autonomous driving benchmarks without outlier exposure, yet its deployment characteristics on embedded hardware have not been systematically characterized. More generally, to the best of our knowledge, no published work provides a systematic latency and energy characterization of anomaly segmentation methods on embedded automotive or railway hardware. This work fills this gap by reformulating PixOOD into a deployment-ready pipeline and benchmarking it on representative embedded and desktop platforms, across both the autonomous driving and railway domains.

\section{Background} \label{sec:background}

This section describes the datasets and the evaluation protocols adopted in this work, also providing a brief overview of the PixOOD architecture, used as the baseline approach.

\subsection{Datasets and evaluation protocols} \label{subsec:datasets}

Standard anomaly segmentation benchmarks for autonomous driving, such as LostAndFound~\cite{LaF}, Fishyscapes~\cite{blum2021fishyscapes}, and SegmentMeIfYouCan (SMIYC)~\cite{smiyc}, adopt AP ($\uparrow$ higher is better) and FPR95 ($\downarrow$ lower is better) as primary evaluation metrics.
These metrics are computed only within a road-region mask: pixels outside the road and sidewalk (e.g., sky, buildings, and vegetation) receive an \emph{ignore} label and are excluded from all calculations. While this facilitates comparisons among autonomous driving methods, it conceals detections that would trigger spurious alarms when the full image is processed.

Among the available autonomous driving benchmarks, LaF is selected as the primary evaluation set for this work because, unlike Fishyscapes and SMIYC, it provides complete pixel-level annotations for all objects in the test set, which allows evaluating the full frame rather than only the road-region mask.

\noindent\textbf{LostAndFound} contains 1203 evaluation frames at $2048 \times 1024$ pixels showing small road obstacles. Because LaF provides complete pixel-level annotations, the ignore labels can be removed to define a full-frame protocol that reflects deployment conditions, in which the entire image is processed. In this work, LaF is therefore evaluated under two protocols: the \textit{standard} road-region protocol, for comparability with published rankings, and the \textit{full-frame} protocol, which reveals the off-road behavior of the model (Section~\ref{sec:experimental}).

\begin{figure}[t!]
  \centering
  \begin{minipage}[t]{0.49\textwidth}
    \centering
    \includegraphics[trim=0cm 0cm 0cm 2.4cm, clip, width=\textwidth]{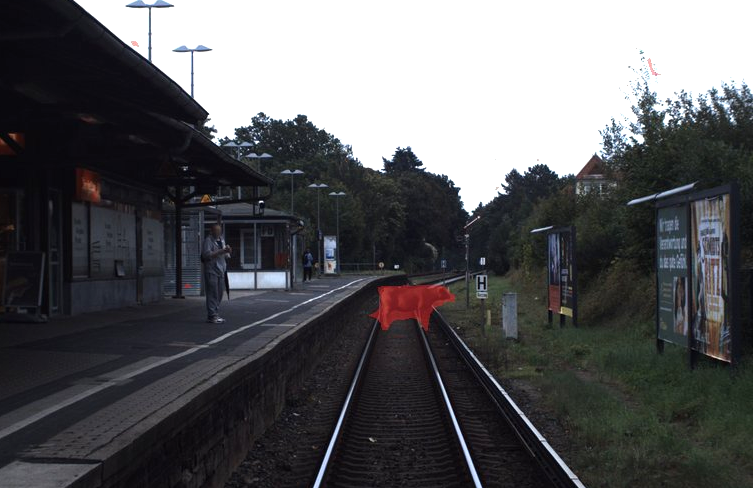}
  \end{minipage}
  \hfill
  \begin{minipage}[t]{0.49\textwidth}
    \centering
    \includegraphics[trim=4cm 0cm 0cm 0cm, clip, width=\textwidth]{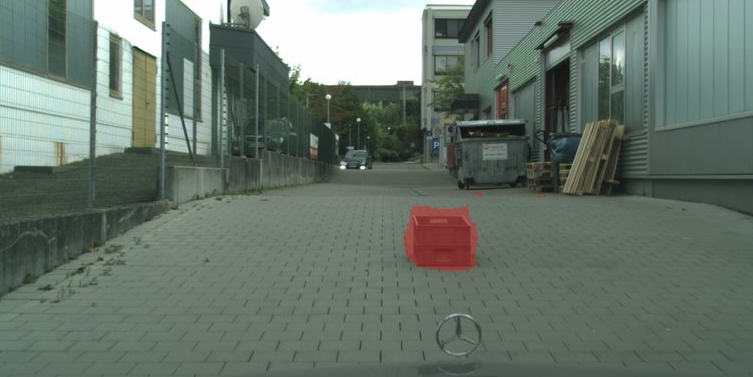}
  \end{minipage}
  \caption{Two examples of anomaly segmentation. Red overlays mark the pixels flagged as anomalous. Left: an OSDaR-AR railway scene, where a cow placed on the tracks is correctly detected as an out-of-distribution obstacle. Right: a LostAndFound road scene, where a small obstacle on the road surface is flagged as an anomaly.
  }
  \label{fig:qualitative}
\end{figure}

\noindent\textbf{OSDaR-AR}~\cite{nesti2026osdar} augments real railway scenes with synthetic out-of-distribution obstacles through augmented reality, providing 1800 frames at $2464 \times 1600$\,px. The obstacles, overlaid as the only annotated objects, make the ground truth binary and full-frame, with no ignore labels. OSDaR-AR is therefore evaluated under the full-frame protocol only. Figure~\ref{fig:qualitative} shows qualitative examples of the accelerated PixOOD pipeline on both datasets.

\subsection{PixOOD architecture}

\begin{figure}[t!]
  \centering
  \includegraphics[width=\textwidth, trim={0 4cm 0 4cm}, clip]{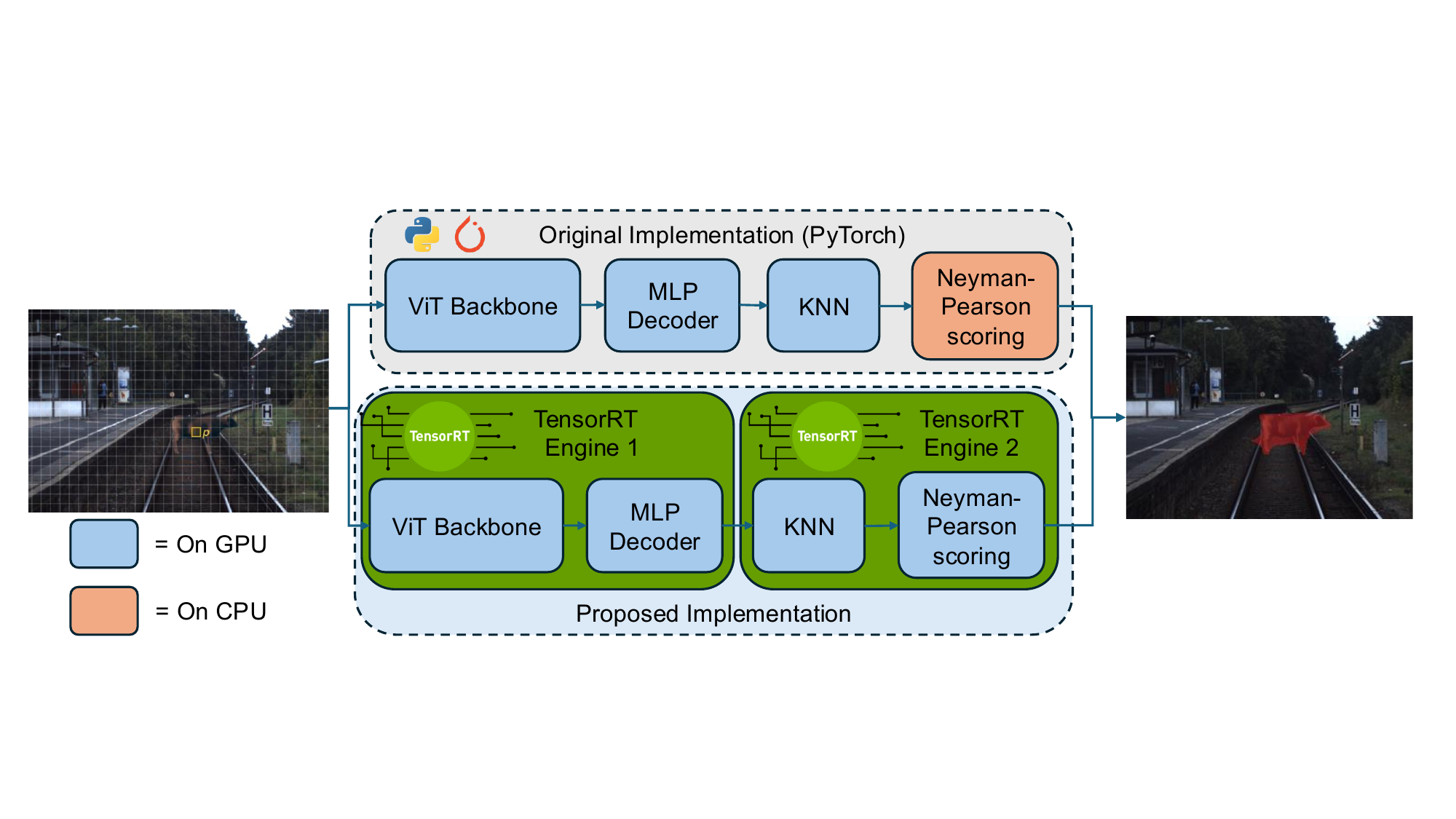}
  \caption{PixOOD processing pipeline. \textit{Top}: the original PyTorch implementation~\cite{vojir2024pixood}, in which the ViT backbone, MLP decoder, and KNN prototype module run on the GPU (blue), while the Neyman--Pearson scoring stage runs on the CPU (orange). \textit{Bottom}: the proposed accelerated implementation, organized into two TensorRT engines that run entirely on the GPU.}
  \label{fig:pixood_architecture}
\end{figure}

The PixOOD processing pipeline is summarized in Figure~\ref{fig:pixood_architecture} and consists of a frozen ViT backbone, an MLP decoder and a K-Nearest Neighbors (KNN) prototype module, both applied to each patch embedding $z$ to produce its two-component vector $z_c$, and a Neyman--Pearson scoring stage based on the likelihood ratio. Each component is described in the following paragraphs.

\noindent\textbf{ViT backbone.} PixOOD uses DINOv2 as a feature extractor backbone, whose weights remain fixed after pre-training. The input image is divided into non-overlapping patches whose size depends on the specific backbone variant (14$\times$14\,px for DINOv2 and 16$\times$16\,px for the more recent DINOv3), and each patch is mapped to a $D$-dimensional embedding vector $z$.

\noindent\textbf{MLP decoder.} A four-layer Multi-Layer Perceptron (MLP) projects each patch embedding $z$ into a class-specific logit $f_c(z)$ for each class $c$ of the $C$ semantic classes of the training distribution.

\noindent\textbf{Prototype condensation and KNN.} During training, a differentiable soft $k$-means algorithm clusters the patch embeddings into $K = 1000$ prototypes per semantic class, providing a compact representation of the feature space within each class distribution. At inference, a batched KNN module computes the L2 distance $d(z, k_{c}^{\ast})$ from each patch embedding $z$ to its nearest prototype $k_{c}^{\ast}$ of class $c$. Together with the per-class MLP logit $f_c(z)$, this distance forms a 2D vector $z_c$:

\begin{equation}
z_c = \bigl(f_c(z),\, d(z,k_{c}^{\ast})\bigr).
\end{equation}

\noindent\textbf{Neyman--Pearson score.} For each semantic class $c$, a two-dimensional Gaussian distribution $\mathcal{N}(\mu_c, \Sigma_c)$ is fitted on the coordinate vectors $z_c$ of in-distribution patches. The anomaly score follows the generalized Neyman--Pearson (N-P) likelihood ratio $r(z_c)$ between the fitted in-distribution density and a reference out-of-distribution density, the optimal statistic for separating in-distribution from out-of-distribution patches~\cite{vojir2024pixood}. To obtain a calibrated score, the ratio is mapped through its cumulative distribution function (CDF) over the in-distribution data of class $c$, denoted $F_c$. This function returns the fraction of in-distribution patches whose ratio does not exceed $r(z_c)$, a calibrated value in $[0,1]$ that is high for patches resembling class $c$ and low otherwise. $F_c$ has no closed form and is therefore precomputed as a lookup table, which at inference is evaluated by interpolation at the ratio of each patch. The final per-pixel score $s(x)$ at image location $x$ is:

\begin{equation}
s(x) = 1 - \max_{c} F_c\bigl(r(z_c(x))\bigr).
\end{equation}

\noindent Intuitively, a patch is flagged as anomalous when its coordinate $z_c$ lies in a low-probability region of every in-distribution class density, i.e., when no known class can plausibly explain it.

\section{Proposed optimizations} \label{sec:method}

This section presents two optimizations that transform PixOOD into a deployment-ready pipeline: a GPU-accelerated N-P scoring reformulation (Section~\ref{subsec:accelerated}) and a TensorRT-based deployment (Section~\ref{subsec:hardware}).

Figure~\ref{fig:pixood_architecture} contrasts the proposed TensorRT-based implementation with the original PyTorch-based baseline.

The original PixOOD implementation exhibits a severe CPU bottleneck in the Neyman--Pearson scoring stage. The CDF lookup is performed on the CPU with SciPy~\cite{vojir2024pixood} in double precision floating point (float64), requiring a round-trip data transfer per frame: the patch embeddings computed on the GPU are transferred to host memory, where SciPy evaluates the multivariate normal density and performs 1D interpolation on the precomputed CDF tables sequentially for each of the $C$ classes. The resulting anomaly scores are then transferred back to the GPU. On the Jetson AGX Orin at 640\,px input resolution, this round-trip alone dominates the 234.7\,ms per-frame inference time, yielding 4.26\,FPS. The GPU reimplementation described in Section~\ref{subsec:accelerated} eliminates this round-trip by running the scoring stage on the GPU.

\subsection{GPU-accelerated Neyman--Pearson scoring} \label{subsec:accelerated}

The original scoring implementation exhibits three crucial bottlenecks: (i) the CPU round-trip described above, (ii) a multivariate normal probability density computation performed in float64 and serialized over the $C$ semantic classes (i.e., one class evaluated at a time, with no parallel exploitation across classes), and (iii) a CPU-bound CDF interpolation routine that is neither parallelizable nor exportable to ONNX.
The proposed GPU reimplementation addresses all three bottlenecks through two complementary components: (i) a log-space GPU kernel for the multivariate normal density, and (ii) a GPU-native CDF lookup.

\noindent\textbf{Log-space multivariate normal distribution.} To exploit the GPU efficiently, the probability density computation uses single-precision floating point (float32), the native precision of GPU compute cores, rather than the double-precision float64 of the original CPU implementation, since float64 throughput is drastically lower on consumer and embedded GPU hardware. In the original CPU implementation, the density is evaluated in linear space using float64, which avoids underflow thanks to the wider representable range compared to float32. The linear-space density of the per-class 2D Gaussian $\mathcal{N}(\mu_c, \Sigma_c)$ might underflow the float32 representable range when the covariance determinant $|\Sigma_c|$ is small. The proposed GPU kernel therefore reformulates the computation entirely in log-space: $\log p(z_c \mid c) = -\tfrac{1}{2}\bigl(d_M^2 + 2\log(2\pi) + \log|\Sigma_c|\bigr)$, where $d_M^2 = (z_c - \mu_c)^\top \Sigma_c^{-1} (z_c - \mu_c)$ is the squared Mahalanobis distance. Since $\Sigma_c$ is fixed after training and is only $2 \times 2$, both $\Sigma_c^{-1}$ and $\log|\Sigma_c|$ are precomputed offline. The resulting tensors are stored on the GPU, and the log-probability for all $C$ classes is then evaluated in a single batched operation across all patches, with no CPU involvement.

\noindent\textbf{GPU-native CDF lookup.} The original implementation evaluates the one-dimensional CDF lookup table on the CPU. For each class, it first computes the per-patch likelihood ratios from their coordinates $z_c$ and then maps each ratio to a score by piecewise-linear interpolation (\texttt{scipy.interpolate.interp1d}), processing one class at a time. Two GPU variants are considered to replace it. The first variant uses \texttt{torch.searchsorted}, a GPU binary-search operator that replicates the original piecewise-linear CDF interpolation in a single batched kernel over all patches and classes. This approach results in AP and FPR95 values that closely match the ones reported in the original paper. Unfortunately, \texttt{torch.searchsorted} is not part of the ONNX operator set and therefore cannot be compiled into a TensorRT engine in the proposed workflow, so it is retained only as an accuracy reference. The second variant is precomputed offline for deployment, relying on \texttt{grid\_sample}, an operator in the ONNX operator set allowing direct compilation into a TensorRT engine. For each class $c$, the complete scoring sequence of Section~\ref{sec:background}, from the evaluation of $p(z_c \mid c)$ to the calibrated score $F_c(r(z_c))$, is evaluated over a uniform $200 \times 200$ grid of $z_c$ coordinates, and the resulting score maps, one per class, are stored on the GPU as static weights. At inference, the scoring sequence is not re-evaluated. Since the maps are defined on a discrete grid while $z_c$ takes continuous values, the score is obtained by bilinear interpolation among the four surrounding grid nodes via \texttt{grid\_sample} on the GPU. The choice of $200 \times 200$ nodes per class balances accuracy and memory, keeping the deviation from the reference negligible while the score maps fit entirely in the GPU cache.

\noindent\textbf{Latency and accuracy.} Running this scoring stage entirely on the GPU removes the per-frame CPU round-trip. Using a PyTorch implementation at 640\,px input resolution, this lowers the per-frame latency from 234.7\,ms to 62.1\,ms on the Jetson AGX Orin (4.26\,FPS to 16.1\,FPS, a $3.78\times$ speedup) and from 106.0\,ms to 22.8\,ms on the desktop RTX~4060 (9.4\,FPS to 44.0\,FPS, a $4.66\times$ speedup), with less than 0.15\,\% AP deviation from the original implementation. Further latency improvements are provided by leveraging TensorRT.

\subsection{Hardware-optimized deployment} \label{subsec:hardware}

The GPU N-P reformulation of Section~\ref{subsec:accelerated} achieves a throughput of 16.1\,FPS on the Jetson AGX Orin, which is still below the 25\,FPS target. Closing this gap requires compiling the full pipeline into TensorRT engines~\cite{aromolo2025real}, rather than leveraging PyTorch for GPU inference. The inference graph is split into two engines executed in sequence: the first contains the ViT backbone and the MLP decoder, the second the KNN module and the \texttt{grid\_sample} variant of the Neyman--Pearson scoring stage.

\noindent\textbf{ONNX export and attention fix.} Both engines are produced from ONNX exports of their source PyTorch modules: the backbone with the MLP decoder forms one graph, and the KNN with the \texttt{grid\_sample} N-P scoring stage forms the other. The backbone export requires an additional fix to ensure attention stability under TensorRT compilation. DINOv2 and DINOv3 implementations use fused scaled dot-product attention (SDPA). When this fused kernel is compiled into TensorRT and executed in FP16 on the Jetson AGX Orin, it becomes numerically unstable and collapses the attention output to a near-zero cosine similarity with respect to the PyTorch reference, a failure mode also reported by Turkcan~\cite{turkcan2026detectrealtimesingleprompt}. This instability does not arise on the desktop RTX~4060 GPU. To address it, the proposed implementation replaces the fused SDPA with an explicit, non-fused implementation before export, preventing TensorRT from selecting the unstable fused kernel. The fix is nonetheless applied to all model architectures on both platforms to keep a single implementation in the evaluation.

\noindent\textbf{Desktop deployment (RTX~4060).} The standard ONNX-to-TensorRT workflow is used. DINOv2 is compiled in half-precision floating-point (FP16) without numerical issues. In contrast, DINOv3 relies on rotary positional embeddings, which PyTorch exports as conditional \texttt{If} nodes. Since the TensorRT compilation of these nodes is numerically unstable in FP16, DINOv3 is deployed in FP32.

\noindent\textbf{Embedded deployment (Jetson AGX Orin).} On the Orin, the ONNX Runtime distribution available for this platform is linked against cuDNN~8, used for accelerating neural operations on GPU, whereas JetPack~6 (the NVIDIA system software stack for Jetson platforms) ships with cuDNN~9. Since the two versions are not binary-compatible, the TensorRT engines are compiled from ONNX and executed through the native TensorRT Python API instead. On both platforms, the deployed scoring stage uses the \texttt{grid\_sample} variant of Section~\ref{subsec:accelerated}, the only one of the two GPU variants that can be exported to ONNX.

\section{Experimental results} \label{sec:experimental}

This section reports empirical results covering detection accuracy under different evaluation protocols on the autonomous driving domain (LostAndFound) and the railway domain (OSDaR-AR), real-time performance on two hardware platforms, and energy consumption on both platforms.

\subsection{Experimental setup} \label{subsec:hwsetup}

The system is evaluated with four trained model variants, all using the ViT-S backbone: DINOv2 and DINOv3, each trained on Cityscapes~\cite{cordts2016cityscapes} (19 urban semantic classes) and RailSem19~\cite{9025646} (19 railway semantic classes), so $C = 19$ for both. These variants are denoted DINOv2-CS, DINOv2-RS, DINOv3-CS, and DINOv3-RS, where CS and RS indicate the Cityscapes and RailSem19 training sets, respectively. Throughout, the resolution label denotes the largest image dimension after resizing. Each side is then rounded down to the nearest multiple of the backbone patch size, as required by the ViT, and the exact input dimensions therefore depend on the backbone and on the image aspect ratio. LostAndFound is evaluated at its native resolution ($2048 \times 1024$\,px) under both the standard road-region protocol and the full-frame protocol described in Section~\ref{subsec:datasets}. OSDaR-AR is evaluated exclusively under the full-frame protocol at three resolutions: 1792\,px (highest evaluated resolution), 896\,px (intermediate accuracy-throughput trade-off), and 640\,px (embedded deployment target, primary configuration on the Jetson AGX Orin).

Hardware validation is conducted on two platforms: (i) a desktop workstation equipped with an NVIDIA RTX~4060 GPU (8\,GB GDDR6), an Intel Core i7-14700F CPU, and 32\,GB of DDR5 system RAM; and (ii) an NVIDIA Jetson AGX Orin (64\,GB unified LPDDR5, 12-core ARM Cortex-A78AE CPU) operating in the MAXN power mode, which enables full SoC performance. Per-frame inference latency is measured using CUDA synchronization events, averaged over 2000 inference iterations. Image loading and normalization are not included in the reported time.

\subsection{Evaluation protocol and detection accuracy}

\begin{table}[t!]
\centering
\caption{LostAndFound detection accuracy at native resolution ($2048 \times 1024$\,px). The full-frame protocol removes all ignore labels. All models are trained on Cityscapes. The best value of each metric within each protocol is shown in bold.}
\label{tab:laf_results}
\begin{tabular}{|l|c|c|c|c|}
\hline
\multirow{2}{*}{\textbf{Model}} & \multicolumn{2}{c|}{\textbf{Road Region}} & \multicolumn{2}{c|}{\textbf{Full Frame}} \\
\cline{2-5}
 & AP $\uparrow$ & FPR95 $\downarrow$ & AP $\uparrow$ & FPR95 $\downarrow$ \\ \hline
DINOv2-CS & \textbf{87.46} & \phantom{0}\textbf{1.81} & \phantom{0}0.20 & 70.49 \\ \hline
DINOv3-CS & 78.86 & 14.61 & \phantom{0}\textbf{0.31} & \textbf{62.11} \\ \hline
\end{tabular}
\end{table}

Table~\ref{tab:laf_results} presents detection accuracy on LostAndFound under both evaluation protocols. Under the standard road-region protocol, DINOv2-CS achieves 87.46\,\% AP, consistent with the published PixOOD baseline that uses the larger ViT-L backbone~\cite{shoeb2025ood}. This confirms that the lighter ViT-S backbone adopted in this work preserves detection accuracy, and it is therefore used in all subsequent experiments. DINOv3-CS achieves lower road-region accuracy (78.86\,\% AP vs.\ 87.46\,\% AP), likely because its larger 16$\times$16\,px patches reduce spatial granularity at the same input resolution.

Under the full-frame protocol, the AP of both variants collapses below 1\,\%. 
This result stems entirely from the off-road regions that the road-region protocol excludes but that a deployed system must process, where the models produce numerous false positives. The standard protocol is therefore optimistic with respect to real deployment, motivating the use of genuinely full-frame benchmarks such as OSDaR-AR.

\begin{table}[t!]
\centering
\caption{Detection accuracy on OSDaR-AR at three evaluation resolutions. All models are trained on RailSem19. The best value of each metric at each resolution is shown in bold.}
\label{tab:osdar_results}
\begin{tabular}{|l|c|c|c|c|c|c|}
\hline
\multirow{2}{*}{\textbf{Model}} & \multicolumn{2}{c|}{\textbf{1792\,px}} & \multicolumn{2}{c|}{\textbf{896\,px}} & \multicolumn{2}{c|}{\textbf{640\,px}} \\
\cline{2-7}
 & AP $\uparrow$ & FPR95 $\downarrow$ & AP $\uparrow$ & FPR95 $\downarrow$ & AP $\uparrow$ & FPR95 $\downarrow$ \\ \hline
DINOv2-RS & 43.62 & 30.07 & \textbf{46.77} & \textbf{31.63} & \textbf{39.51} & \textbf{37.43} \\ \hline
DINOv3-RS & \textbf{56.75} & \textbf{29.77} & 24.88 & 45.63 & 10.60 & 56.42 \\ \hline
\end{tabular}
\end{table}

Table~\ref{tab:osdar_results} reports OSDaR-AR accuracy across three resolutions. DINOv3-RS achieves the best accuracy at 1792\,px (56.75\,\% AP, 29.77\,\% FPR95), while DINOv2-RS peaks at 896\,px (46.77\,\% AP, 31.63\,\% FPR95), slightly outperforming its 1792\,px result (43.62\,\% AP) due to increased patch density in the anomaly region at the intermediate resolution. DINOv3-RS degrades significantly below 1792\,px since its larger patch size generates fewer patches at lower resolutions, reducing the spatial granularity available to detect the small anomaly stickers.

\subsection{Real-time performance and energy efficiency}

\begin{table}[t!]
\centering
\setlength{\tabcolsep}{3pt}
\caption{Original baseline vs.\ TensorRT pipeline on the desktop NVIDIA RTX~4060 (\emph{4060}) and the NVIDIA
Jetson AGX Orin (\emph{Orin}, MAXN power mode), averaged over 2000 inference instances. \emph{Baseline}
columns are the original SciPy CPU scoring stage; \emph{TensorRT} columns are the deployed pipeline, at the
same precision. The baseline energy does not include the CPU scoring cost and is therefore a lower bound.
For each platform, the best value in each column is shown in bold.}
\label{tab:pipeline_perf}
\resizebox{\columnwidth}{!}{%
\begin{tabular}{|l|l|c|c|c|c|c|c|}
\hline
\multirow{2}{*}{\textbf{Platf.}} & \multirow{2}{*}{\textbf{Model}} & \multirow{2}{*}{\textbf{Res.}} & \multirow{2}{*}{\textbf{Prec.}} & \multicolumn{2}{c|}{\textbf{Speed (ms / FPS)}} & \multicolumn{2}{c|}{\textbf{Energy (J)}} \\
\cline{5-8}
 & & & & \textbf{Baseline} & \textbf{TensorRT} & \textbf{Baseline} & \textbf{TensorRT} \\ \hline
\multirow{6}{*}{4060} & DINOv2 & 640\,px & FP16 & \textbf{109.4 / \phantom{0}9.1} & \textbf{5.51 / 181.6} & \textbf{\phantom{0}6.23} & \textbf{0.56} \\
                     & DINOv2 & 896\,px & FP16 & 205.7 / \phantom{0}4.9 & 12.44 / \phantom{0}80.4 & 11.83 & 1.36 \\
                     & DINOv2 & 640\,px & FP32 & 115.5 / \phantom{0}8.7 & 11.93 / \phantom{0}83.8 & \phantom{0}7.15 & 1.31 \\
                     & DINOv2 & 896\,px & FP32 & 232.4 / \phantom{0}4.3 & 37.97 / \phantom{0}26.3 & 14.88 & 4.30 \\
                     & DINOv3 & 640\,px & FP32 & 119.3 / \phantom{0}8.4 & \phantom{0}9.38 / 106.6 & \phantom{0}7.16 & 1.02 \\
                     & DINOv3 & 896\,px & FP32 & 225.6 / \phantom{0}4.4 & 26.10 / \phantom{0}38.3 & 13.82 & 2.93 \\ \hline
\multirow{6}{*}{Orin} & DINOv2 & 640\,px & FP16 & \textbf{240.2 / \phantom{0}4.2} & \textbf{13.34 / 75.1} & \textbf{1.80} & \textbf{0.44} \\
                      & DINOv2 & 896\,px & FP16 & 447.6 / \phantom{0}2.2 & 28.16 / 35.5 & 3.37 & 1.00 \\
                      & DINOv2 & 640\,px & FP32 & 260.4 / \phantom{0}3.8 & 25.60 / 39.1 & 2.01 & 0.89 \\
                      & DINOv2 & 896\,px & FP32 & 504.4 / \phantom{0}2.0 & 70.98 / 14.1 & 4.13 & 2.63 \\
                      & DINOv3 & 640\,px & FP32 & 306.4 / \phantom{0}3.3 & 21.73 / 46.0 & 2.52 & 0.71 \\
                      & DINOv3 & 896\,px & FP32 & 555.0 / \phantom{0}1.8 & 50.85 / 19.7 & 4.65 & 1.83 \\ \hline
\end{tabular}%
}
\end{table}

The full proposed pipeline is compiled with TensorRT, as described in Section~\ref{subsec:hardware}. Table~\ref{tab:pipeline_perf} reports the per-frame latency, throughput, and energy on both platforms.
For comparison, Table~\ref{tab:pipeline_perf} also reports the latency and energy of the original baseline, over which the deployed pipeline is $6$--$20\times$ faster and consumes less energy per frame at matching precision.
On the Jetson AGX Orin, the DINOv2-RS FP16 pipeline reaches 75.1\,FPS at the 640\,px label and 35.5\,FPS at 896\,px, comfortably exceeding the 25\,FPS safety target, while the DINOv3-RS FP32 pipeline reaches 46.0\,FPS at 640\,px. On the RTX~4060, the same two configurations reach 181.6\,FPS and 106.6\,FPS at 640\,px, respectively. As discussed in Section~\ref{subsec:hardware}, DINOv3 runs in FP32 on both platforms because of the FP16 instability of its rotary positional embeddings. Table~\ref{tab:pipeline_perf} also reports DINOv2-RS in FP32 as a precision reference. At 640\,px and equal precision, DINOv3-RS is faster than DINOv2-RS on both platforms (21.73\,ms vs. 25.60\,ms on the Orin, 9.38\,ms vs. 11.93\,ms on the RTX~4060), but with lower detection accuracy reported in Table~\ref{tab:osdar_results}. The FP16 backbone provides a $1.92\times$ end-to-end speedup over its FP32 counterpart on the Orin at 640\,px (13.34\,ms vs. 25.60\,ms). On the Jetson AGX Orin, energy is characterized through the on-chip \texttt{tegrastats} monitor, which samples the SoC GPU power rail (\texttt{VDD\_GPU\_SOC}) at 100\,ms intervals. Energy per frame is then obtained by multiplying the mean power draw by the per-frame inference time. The DINOv2-RS FP16 configuration consumes 0.44\,J per frame at the 640\,px label (32.94\,W mean draw), rising to 1.00\,J at 896\,px. On the RTX~4060, the available sensors report the total board power, which includes the GPU die, the GDDR6 memory, and the on-board voltage regulators. The corresponding values (0.56 and 1.36\,J for DINOv2-RS FP16) are therefore board-level. The Orin energy is measured at the chip-level GPU power rail, whereas the RTX~4060 energy reflects the total board power and is therefore not directly comparable.

\section{Conclusions} \label{sec:conclusions}

This work presented a systematic benchmark of PixOOD on both desktop and embedded platforms, demonstrating its viability for real-time deployment across two domains, autonomous driving and railway operations. The GPU reimplementation of the Neyman--Pearson scoring stage achieved a $3.78\times$ end-to-end speedup on the Jetson AGX Orin, from 4.26\,FPS to 16.1\,FPS. Combined with platform-specific TensorRT compilation, the full pipeline reaches up to 182\,FPS on the RTX~4060 and 75\,FPS on the Orin (DINOv2-RS FP16 at the 640\,px label, 0.44\,J per frame), well above the 25\,FPS real-time target on both platforms. The dual-protocol evaluation of LostAndFound further showed that the standard road-region protocol is optimistic with respect to real deployment conditions, motivating the adoption of full-frame benchmarks such as OSDaR-AR.

To the best of our knowledge, the existing literature on anomaly segmentation focuses primarily on detection accuracy for autonomous driving, leaving the railway domain and the real-time deployment characterization largely unexplored. Future work targets the extension of the latency and energy characterization to additional state-of-the-art anomaly segmentation methods, the evaluation of the pipeline on a broader set of embedded platforms, and the exploration of multimodal anomaly segmentation through camera-LiDAR fusion using the calibrated point clouds provided by OSDaR-AR.

\bibliographystyle{splncs04}
\bibliography{references}

\end{document}